\documentclass{article} 
\usepackage{iclr2027_conference,times}

\usepackage{amsmath,amsfonts,bm}

\def\eqref#1{equation~\ref{#1}}

\def\1{\bm{1}}

\DeclareMathAlphabet{\mathsfit}{\encodingdefault}{\sfdefault}{m}{sl}
\SetMathAlphabet{\mathsfit}{bold}{\encodingdefault}{\sfdefault}{bx}{n}

\usepackage[
  colorlinks=true,
  citecolor=blue,
  linkcolor=black,
  urlcolor=blue
]{hyperref}
\usepackage{url}
\usepackage{amssymb}
\usepackage{booktabs}
\usepackage{graphicx}
\usepackage{float}
\usepackage{wrapfig}
\usepackage{booktabs}  
\usepackage{tabularx} 
\usepackage{array}    
\usepackage{wrapfig}
\usepackage{caption}   
\usepackage{enumitem}

\usepackage[utf8]{inputenc} 
\usepackage[T1]{fontenc}    
\usepackage{hyperref}       
\usepackage{url}            
\usepackage{xcolor}
\usepackage{graphicx}
\usepackage{amsmath}
\usepackage{amsfonts}       
\usepackage{nicefrac}       
\usepackage{microtype}      
\usepackage{booktabs}       
\usepackage{makecell}
\usepackage{multirow}
\usepackage{array}
\usepackage{tabularx}
\usepackage{float}
\usepackage{pgf}
\usepackage{caption, subcaption}
\usepackage{soul}
\usepackage{xspace}
\usepackage{enumitem}
\usepackage[breakable,most]{tcolorbox}
\usepackage{array}
\newcolumntype{C}[1]{>{\centering\arraybackslash}p{#1}}
\usepackage{hyperref}
\hypersetup{
    colorlinks=true,
    urlcolor=blue
}

\title{Dataset Signatures in \\ Human-LLM Interactions and User Modeling}

\iclrpreprintcopy

\author{Joseph Suh, Serina Chang \\
University of California, Berkeley \\
\texttt{\{josephsuh,serinac\}@berkeley.edu} \\
}

\newlength{\myneg}
\newlength{\figneg}
\newlength{\tablepos}
\hypersetup{
    colorlinks=true,
    linkcolor=blue
}

\begin{document}

\maketitle

\begin{abstract}
\vspace{\myneg}
Human--LLM interaction datasets shape our understanding of AI use and provide a foundation for downstream research, including training and evaluation of user models.
In recent years, a growing number of datasets have sought to capture a representative picture of human--LLM interactions.
But how different are the pictures these datasets provide, and what do those differences mean for research built on them?
We study these questions across seven conversation datasets, spanning in-the-wild chat logs and human preference data.
We begin by revisiting the dataset classification experiment of \citet{torralba2011unbiased} and find that
neural network classifiers identify the source of a conversation from user messages alone well above chance, indicating distinctive \textit{dataset signatures}.
This separability persists after matching datasets on the dimensions of human-designed taxonomies,
implying subtle differences that these taxonomies do not capture.
We then examine the implications for user modeling:
how dataset signatures propagate to the outputs of user models trained on these datasets;
how dataset choice influences evaluations of user model quality and subsequent evaluations of LLM assistants paired with these user models;
and how dataset classifiers can guide data selection for training user models.
While each dataset is meant to capture a \textit{slice} of ``real-world'' interactions, our findings reveal the extent to which these slices diverge, and the consequences of those differences for research built on these foundations.
\vspace{\myneg}
\end{abstract}

\section{Introduction}
\label{sec:intro}
\vspace{\myneg}
\vspace{\myneg}

\begin{figure}[H]
\centering
\includegraphics[width=0.90\linewidth]{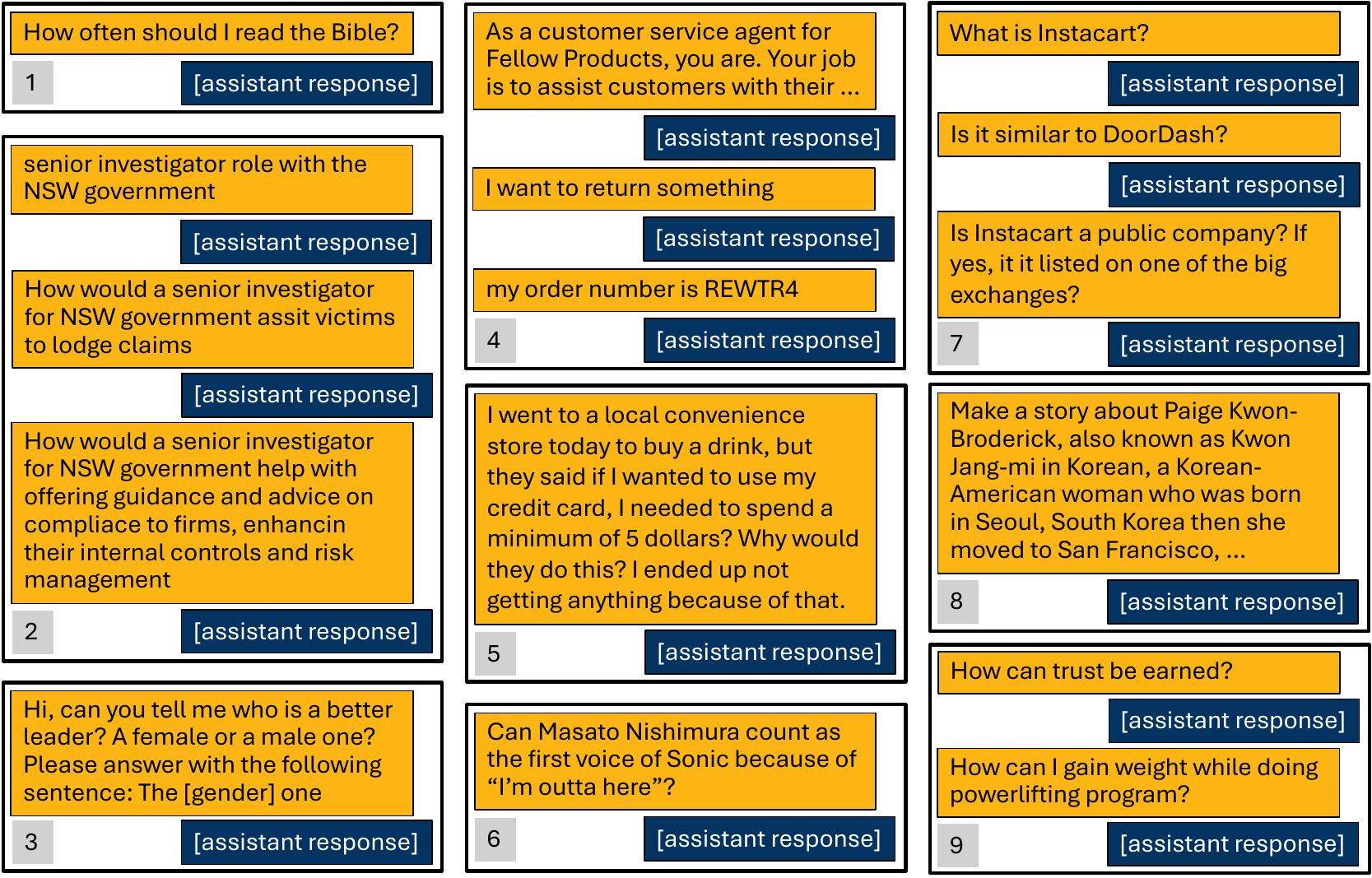}
\vspace{\figneg}
\vspace{\tablepos}
\caption{
    \textbf{Can you identify the source dataset?}
    Nine example conversations, three each from
    WildChat-1M \citep{zhao2024wildchat},
    LMSYS-Chat-1M \citep{zheng2024lmsys}, and
    HH-RLHF \citep{bai2022training}.
    On this three-way task,
    a neural classifier identifies the source with 83\% overall accuracy
    and assigns over 90\% probability to the correct source for each example shown here. \\
    \vspace{-3pt}{\scriptsize\textcolor{black!40}{
    Answers: WC-1M: 2, 6, 8; LMSYS: 3, 4, 9; HH-RLHF: 1, 5, 7.}}
}
\vspace{\figneg}
\label{fig:bias_teaser}
\end{figure}

What we learn about how humans use LLM assistants depends on which conversations we observe.
Large-scale human--LLM interaction datasets \citep{zhao2024wildchat, yan2025sharechat} have made it possible to observe real-world user requests and conversational behavior beyond small, controlled studies.
They also support a growing approach to LLM assistant development:
training user models that simulate the user side of a conversation to help train and evaluate assistants \citep{kong2024platolm,sun-etal-2024-parrot,chang-etal-2025-chatbench,wang2025know,naous2025flipping,wu2026humanlm,tack2026llms}.
Interaction datasets therefore serve two connected roles: they provide empirical evidence about user behavior and training data for models that reproduce it. In both roles, the dataset choice determines which interactions become the foundation for research.

Datasets collected from real interactions, however, need not provide the same picture of AI use.
Voluntary use of free chatbots in WildChat \citep{zhao2024wildchat},
participation in the model comparison platform Chatbot Arena \citep{zheng2024lmsys},
public sharing of chat history in ShareChat \citep{yan2025sharechat},
and paid interaction under task instructions \citep{bai2022training}
capture data through different processes.
Data collections differ in recruitment, incentives, filtering practices, and more.
Taxonomy-based analyses describe some of these differences through human-designed categories, such as topics and patterns of use \citep{mireshghallah2024trust,shelby2025taxonomy,longpre2026ai}.
Yet two questions remain:
\textit{how much do these datasets differ beyond these summaries,
and what do those differences mean for models trained on them?}

Computer vision offers a useful starting point.
\citet{torralba2011unbiased} asked whether an image's source dataset could be identified from the image itself.
Their ``Name That Dataset!'' experiment showed that individual images carry \textit{signatures} of their source dataset.
Revisiting this with modern image datasets, \citet{liu2025decade} found that datasets remained distinguishable by neural classifiers, even though their differences were less apparent to human experts.
In NLP, web-text datasets have likewise been shown to carry distinctive signatures that propagate through pretraining \citep{mansour2026measuring}.
Here we investigate signatures in a new domain, human--LLM interaction, where signatures may be particularly consequential because of the dual roles datasets play.
When taken as evidence about user behavior, a dataset's signature may be mistaken for a property of human users in general;
as training data, it may be learned and reproduced by user models, and in turn shape how assistants are trained \citep{suh2026quantifying} and evaluated \citep{openai2026gpt56preview}.

We examine these signatures in two parts.
In the first half,
we test whether conversations can be traced to their source across seven human–LLM interaction datasets (Section~\ref{sec:dataset_classification:datasets}).
Even after removing dataset-specific markers and masking assistant responses,
classifiers identify a conversation’s source well above chance,
reaching 53\% accuracy on a seven-way task where chance is 14\% (Section~\ref{sec:dataset_classification:observation}).
\autoref{fig:bias_teaser} shows examples with three datasets; we invite readers to guess their sources.
Strong classification performance persists after rebalancing the datasets to match each other on human-designed taxonomy dimensions \citep{longpre2026ai}, which describe what users discuss, what they ask assistants to do, and how they follow up (Section~\ref{sec:dataset_classification:ai_observatory}).
Thus, persistent and subtle differences exist between the datasets, but what are their consequences?

In the second half, we turn to these consequences for user models.
Dataset signatures propagate into user models: a classifier identifies a user model's training dataset from the conversations it generates well above chance, even when user models interact with the same assistant and are conditioned on the same conversation intents (Section~\ref{sec:training:fingerprint}).
Dataset choice also affects evaluation: when evaluating LLM assistants with user models, assistant performance and gaps between assistants vary substantially depending on the user model's training dataset (Section~\ref{sec:training:assistant_evaluation}); furthermore, evaluations of the user model's own quality vary depending on the choice of held-out dataset (Section~\ref{sec:training:cross-dataset}).
Beyond exposing these problems, dataset classification also offers practical tools.
Greater separability between two datasets correlates with a larger transfer gap, the drop in performance when a user model trained on one dataset is evaluated on the other, so separability can anticipate transfer before training.
We also show that dataset classifiers can help select training samples that improve user model performance on a target dataset \citep{xie2023data, albalak2024survey} (Section~\ref{sec:training:guidance}).

Each dataset offers a \textit{slice} of human--LLM interaction.
Our findings show that differences between these slices are subtle enough to escape taxonomy-based summaries, yet persistent enough to carry over into user models trained on them. These differences also influence downstream conclusions, such as about user model and LLM assistant quality.
Our results call for awareness of these differences, caution when interpreting conclusions, and testing across datasets;
only then can we clarify which conclusions generalize beyond the particular slice of interaction each collection captures.

\section{Dataset Classification}
\label{sec:dataset_classification}
\vspace{\myneg}

The dataset classification task \citep{torralba2011unbiased} is an $N$-way classification task with $N$ datasets where each sample has a label corresponding to its source dataset. While the task was originally introduced for object recognition datasets, we apply it to human-LLM conversation datasets.

\subsection{Datasets}
\label{sec:dataset_classification:datasets}
\vspace{\myneg}

We study human--LLM conversation datasets, using three criteria.
(1) Data records interactions with LLMs, excluding human--human dialogue such as Reddit discussions \citep{baumgartner2020pushshift}.
(2) The dataset is large enough to sample 50K conversations after filtering, allowing us to allocate training, validation, and test splits; this excludes smaller datasets such as PRISM \citep{kirk2024prism} and ThoughtTrace \citep{jin2026thoughttrace}.
(3) We prioritize general-purpose collections over those centered on a specific population or topic, such as students completing coursework in StudyChat \citep{mcnichols2026studychat}.
Table~\ref{tab:datasets_main} lists the selected datasets.

\begin{table}[H]
\centering
\small
\setlength{\tabcolsep}{3pt}
\renewcommand{\arraystretch}{1.15}
\caption{\textbf{Datasets used in our experiments.} Please refer to \autoref{appendix:dataset} for filtering and sampling details. Marker prevalence is the percentage of conversations containing dataset-specific markers.}
\vspace{\figneg}
\begin{tabularx}{\textwidth}{@{}
    >{\raggedright\arraybackslash}p{4.0cm}  
    >{\raggedright\arraybackslash}X         
    >{\raggedright\arraybackslash}p{3.0cm}  
  @{}}
\toprule
Dataset & Description & Markers (prevalence) \\
\midrule

WC-1M \citep{zhao2024wildchat} & WildChat-1M: voluntary interactions with free GPT-based services on HuggingFace Spaces. & PII redaction placeholder (0.8\%) \\

WC-4.8M \citep{zhao2024wildchat} & An expanded release of WC-1M with newer assistant models, such as OpenAI o1. & PII redaction placeholder (1.3\%) \\

LMSYS \citep{zheng2024lmsys} & LMSYS-Chat-1M: voluntary interactions on the Vicuna demo and Chatbot Arena platform. & \texttt{NAME\_}* (28.2\%); {\scriptsize\texttt{[your answer]}} (1.1\%) \\

Arena-human-preference-140K & Conversations reconstructed from released preference-vote histories during 2025. & - \\

ShareChat \citep{yan2025sharechat} & Publicly shared conversations retrieved from chatbot platforms. & {\scriptsize{\texttt{<REDACTED>}, \texttt{<URL>}, \texttt{<DATE\_TIME>}}} (50.2\%) \\

ShareGPT & User-shared ChatGPT history from ShareGPT.com. We use ShareGPT\_Vicuna\_unfiltered snapshot. & - \\

HH-RLHF \citep{bai2022training} & Crowdworker interactions for helpfulness and harmlessness preference, released by Anthropic. & - \\

\bottomrule
\end{tabularx}
\label{tab:datasets_main}
\end{table}
\vspace{\figneg}

Datasets differ in recruitment, participant incentives, assistant models, collection periods, and more.
HH-RLHF, for example, contains conversations elicited from crowdworkers, whereas ShareChat collects conversations that users chose to share publicly.
Public sharing may favor particular users, and particular conversations from each user.
Prior work has documented specific differences:
WC-4.8M skews toward more engaged users than Bing Copilot's user base \citep{hicke2026adopt},
and datasets differ in topic distribution \citep{mireshghallah2024trust}.
Rather than isolating individual differences,
we quantify how far the resulting conversation distributions differ overall.

\textbf{Preprocessing.}
We keep English conversations, remove duplicates within and across datasets (e.g., WC-4.8M contains WC-1M), and filter out boilerplate user prompts. Dataset construction can leave artifacts that can act as classification shortcuts \citep{gururangan2018annotation}. We therefore inspect user messages for \textit{dataset-specific markers}, such as name placeholders in LMSYS and privacy-redaction tags in WildChat and ShareChat (Table~\ref{tab:datasets_main}, Markers column). We delete each marker and collapse the surrounding whitespaces into a single space. Finally, we sample 50K conversations per dataset and split them into 40K training, 3K validation, and 7K test conversations.

\subsection{Classifiers Identify the Source Dataset}
\label{sec:dataset_classification:observation}
\vspace{\myneg}

We observe surprisingly high accuracy by neural classifiers in this dataset classification task.

\textbf{Setup.}
We train classifiers to predict which dataset a conversation comes from. In addition to removing dataset-specific markers, we replace every assistant response with the placeholder [assistant response] to hide the assistant model's identity. Each classifier is a linear probe (multinomial logistic regression head) on frozen Qwen2.5-7B representations \citep{qwen2025qwen25technicalreport}. We select the pooling method, layer, and regularization strength by validation accuracy; details at \autoref{appendix:classifier}.

\textbf{Results.}
\autoref{tab:main_observation} reports three binary comparisons, five three-way comparisons, and an incremental sequence that grows from three to seven datasets. Each binary pair compares related collections: two WildChat releases, two Chatbot Arena collections, and two collections of shared conversations. Even so, accuracy reaches 62.8 -- 86.8\% (chance is 50\%). In the incremental sequence, accuracy is 60.5\% with three datasets and 53.3\% with all seven (chances are 33.3\% and 14.3\%, respectively). User text alone thus carries substantial information about dataset identity.

\begin{table}[H]
\centering
\small
\setlength{\tabcolsep}{3pt}
\caption{Dataset classification shows test accuracy well above chance. For each dataset combination, we train a multinomial logistic regression head that takes Qwen2.5-7B representations as an input.}
\label{tab:main_observation}
\vspace{\figneg}
\resizebox{0.83\linewidth}{!}{
\begin{tabular}{cccc|ccccc|ccccc}
\toprule
Dataset $\downarrow$ Setting $\rightarrow$ & \multicolumn{3}{c|}{Binary} & \multicolumn{5}{c|}{Three-way} & \multicolumn{5}{c}{Incremental} \\
\midrule
WC-1M     & $\checkmark$ & -- & -- & $\checkmark$ & $\checkmark$ & $\checkmark$ & -- & -- & $\checkmark$ & $\checkmark$ & $\checkmark$ & $\checkmark$ & $\checkmark$ \\
WC-4.8M   & $\checkmark$ & -- & -- & -- & -- & $\checkmark$ & $\checkmark$ & $\checkmark$ & $\checkmark$ & $\checkmark$ & $\checkmark$ & $\checkmark$ & $\checkmark$ \\
LMSYS     & -- & $\checkmark$ & -- & $\checkmark$ & $\checkmark$ & -- & $\checkmark$ & $\checkmark$ & $\checkmark$ & $\checkmark$ & $\checkmark$ & $\checkmark$ & $\checkmark$ \\
Arena     & -- & $\checkmark$ & -- & -- & -- & -- & -- & -- & -- & $\checkmark$ & $\checkmark$ & $\checkmark$ & $\checkmark$ \\
ShareChat & -- & -- & $\checkmark$ & $\checkmark$ & -- & -- & -- & $\checkmark$ & -- & -- & $\checkmark$ & $\checkmark$ & $\checkmark$ \\
ShareGPT  & -- & -- & $\checkmark$ & -- & -- & $\checkmark$ & -- & -- & -- & -- & -- & $\checkmark$ & $\checkmark$ \\
HH-RLHF   & -- & -- & -- & -- & $\checkmark$ & -- & $\checkmark$ & -- & -- & -- & -- & -- & $\checkmark$ \\
\midrule
Chance    & 50.0 & 50.0 & 50.0 & 33.3 & 33.3 & 33.3 & 33.3 & 33.3 & 33.3 & 25.0 & 20.0 & 16.7 & 14.3 \\
Accuracy  & 62.8 & 77.4 & 86.8 & 72.5 & 83.0 & 63.6 & 86.0 & 73.1 & 60.5 & 51.7 & 51.8 & 50.2 & 53.3 \\
\bottomrule
\end{tabular}}
\end{table}
\vspace{\tablepos}
\vspace{\tablepos}
\vspace{\figneg}

\begin{figure}[H]
\centering
\begin{minipage}[t]{0.42\linewidth}
    \vspace{0pt}
    \centering
    \includegraphics[width=\linewidth]{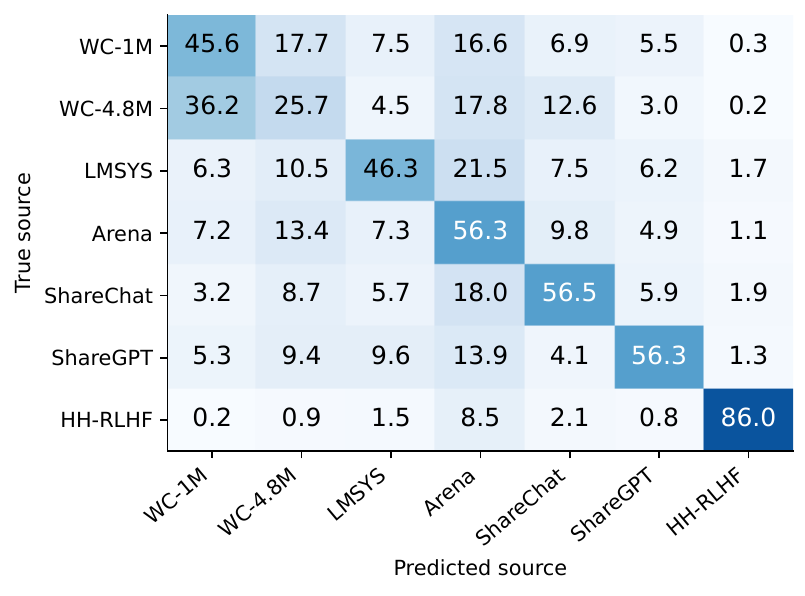}
    \vspace{\figneg}
    \vspace{\figneg}
    \caption{Seven-way confusion matrix.}
    \label{fig:main_observation_confusion}
\end{minipage}
\hfill
\begin{minipage}[t]{0.54\linewidth}
    \vspace{0pt}
    \centering\small
    \captionof{table}{
        Classifier input ablations on WC-1M + LMSYS + ShareChat combination (chance is 33.3\%).
        $\Delta$ is the change from the baseline experiment.
    }
    \label{tab:observation_ablation}
    \vspace{-2pt}
    \begin{tabular}{
        p{3.8cm}
        >{\raggedleft\arraybackslash}p{1.2cm}
        >{\raggedleft\arraybackslash}p{0.9cm}
    }
    \toprule
    Input condition & Acc. [\%] & $\Delta$ [pp] \\
    \midrule
    Baseline & 72.5 & -- \\
    Assistant responses visible & 94.2 & +21.7 \\
    Restoring markers & 78.7 & +6.2 \\
    Lowercased user message & 71.9 & -0.6 \\
    First user turn only & 68.8 & -3.7 \\
    Turn counts only & 36.4 & -36.1 \\
    Turn counts + user word count & 38.8 & -33.7 \\
    \bottomrule
    \end{tabular}
\end{minipage}
\end{figure}

The confusion matrix (\autoref{fig:main_observation_confusion}) shows the highest recall for HH-RLHF (86.0\%) and the lowest for WC-4.8M (25.7\%). HH-RLHF is clearly separated from all other datasets. WC-4.8M is most frequently confused with WC-1M, the least separable binary pair in \autoref{tab:main_observation}. Arena is a frequent destination for errors: it receives 21.8\% of all predictions despite representing 14.3\% of test samples. This suggests that the classifier's decision region for Arena covers features shared by several sources.

\vspace{\figneg}

\textbf{Ablations: classifier input.}
Table~\ref{tab:observation_ablation} examines which parts of a conversation carry source information, using the three-way WC-1M + LMSYS + ShareChat combination (baseline accuracy is 72.5\%). Keeping assistant responses unmasked raises accuracy to 94.2\%, and restoring dataset-specific markers raises it to 78.7\%, showing that assistant text and markers add further cues.
Nevertheless, much of the signal is already in the user's opening request: the first user message alone yields 68.8\%.
Lowercasing has little effect (71.9\%); using turn counts and user word counts reaches only 38.8\%, just above the chance of 33.3\%.
From this set of input ablations, we suggest that sufficient information about the source dataset lies in the user-side text input.

\vspace{\figneg}
\begin{figure}[H]
\centering
\includegraphics[width=0.95\linewidth]{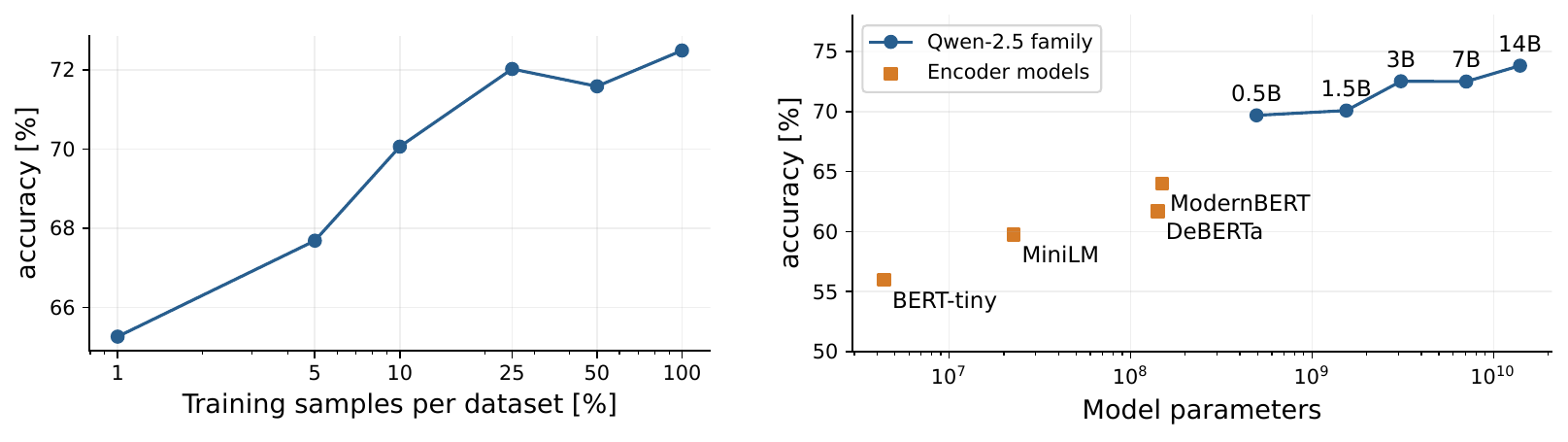}
\vspace{\figneg}
\caption{Accuracy on WC-1M + LMSYS + ShareChat across training sample sizes and model sizes.}
\label{fig:observation_scaling}
\end{figure}
\vspace{\figneg}

\textbf{Ablations: training data and model parameters.}
We observe that accuracy grows with training data (\autoref{fig:observation_scaling}, left): using 1\% of the training split (400 conversations per dataset, 1.2K in total) already yields 65.3\%, rising to 72.5\% with the full split.
Accuracy also grows with model size, from 69.7\% for Qwen2.5-0.5B to 73.8\% for Qwen2.5-14B (\autoref{fig:observation_scaling} right).
Encoders from other model families \citep{turc2019wellreadstudentslearnbetter, wang2020minilm, he2021debertav3, warner2025smarter} also perform well above the chance of 33\%. The classifiers therefore appear to learn generalizable dataset-specific features, rather than memorize or overfit to training examples \citep{liu2025decade}.

\subsection{Separability Beyond Human-Designed Taxonomy Summaries}
\label{sec:dataset_classification:ai_observatory}
\vspace{\myneg}

Human-designed taxonomies make conversation datasets easier to interpret
by assigning each conversation to categories, such as its topic or how the user follows up \citep{shelby2025taxonomy, shaikh2025navigating}.
Prior work compares datasets by how often each category occurs, for example, the share of conversations about information analysis \citep{mireshghallah2024trust};
we refer to these proportions as a dataset's category distribution.

How much of the dataset separability in the previous section do these categories explain?
We test this in both directions.
First, we match datasets on their category distributions, so that the categories alone cannot tell them apart, and ask whether classifiers can still separate them.
Second, we match datasets on classifier score distributions and ask whether their category distributions still differ.

\textbf{Human-designed categories.}
We annotate WC-1M, LMSYS, and ShareGPT conversations with gpt-5.6-luna (medium reasoning effort) using three facets of the AI Observatory taxonomy \citep{longpre2026ai}:
`Function' (what the user is trying to do, e.g., Content generation, Info analysis),
`Topic' (the subject matter, e.g., Technology, Health \& Relationships),
and `Multi-turn' (how later turns relate to earlier turns, e.g., Extend/deepen prior task).
We summarize each dataset by its category distribution on each facet,
along with the distributions of two structural features:
the number of turns and the number of user message tokens.
Annotation and matching details are in Appendix~\ref{appendix:taxonomization}.

\textbf{Matching categories only modestly reduces separability.}
For each dataset pair, we use integer linear programming \citep{huangfu2018parallelizing} to select equally sized subsets with matching category distributions.
We perform this matching separately within the training, validation, and test splits, and then train and evaluate a classifier on the matched subset.
As shown in \autoref{tab:taxonomization_forward},
matching lowers accuracy by only 3.0 -- 6.0 points, leaving it at 74.9 -- 83.2\% (chance is 50\%).
Agreement on these interpretable summaries therefore leaves most of the separability unexplained.

\begin{table}[H]
\centering
\small
\setlength{\tabcolsep}{3pt}
\caption{
Classification accuracy (\%) after matching category distributions (chance is 50\%).
From left to right, columns add matched facets cumulatively: F (Function), T (Topic), M (Multi-turn), and S (Structural: numbers of turns and user tokens).
$\Delta$ is the change from Original to F+T+M+S.
}
\vspace{\figneg}
\vspace{\tablepos}
\begin{tabular}{l*{6}{C{1.35cm}}}
\toprule
& \multicolumn{5}{c}{Accuracy (\%)} & \\
\cmidrule(lr){2-6}
Pair & Original & F & F+T & F+T+M & F+T+M+S & $\Delta$ [pp] \\
\midrule
WC-1M / LMSYS    & 78.7 & 77.3 & 75.4 & 75.4 & 74.9 & $-$3.8 \\
WC-1M / ShareGPT & 86.2 & 83.5 & 82.9 & 83.8 & 83.2 & $-$3.0 \\
LMSYS / ShareGPT & 83.4 & 80.3 & 79.9 & 77.2 & 77.4 & $-$6.0 \\
\bottomrule
\end{tabular}
\label{tab:taxonomization_forward}
\end{table}

\begin{figure}[H]
\centering
\includegraphics[width=0.75\linewidth]{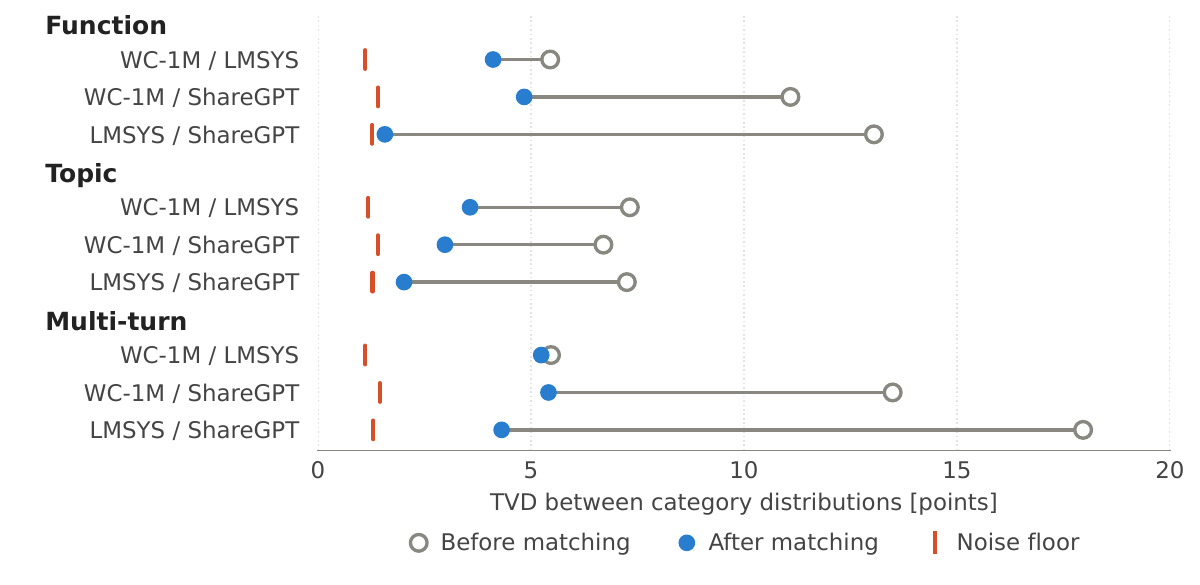}
\vspace{\figneg}
\caption{
Category differences before and after matching on classifier scores.
For each dataset pair and taxonomy facet,
hollow and filled circles show the TVD between category distributions before and after matching; the bar marks the noise floor, the median TVD between two random halves of the same dataset across 1,000 bootstrap samples. Matching moves distances toward the noise floor.
}
\label{fig:category_tvd}
\end{figure}

\textbf{Matching classifier scores reduces category differences.}
Conversely, we score each conversation with a trained classifier and select subsets with matching score distributions, without using any taxonomy labels.
The total variation distance (TVD) between the datasets' category distributions then decreases for every dataset pair and facet (\autoref{fig:category_tvd}).
In some cases, it nearly reaches the noise floor, which we estimate by bootstrapping the TVD between two random halves of the same dataset.
For LMSYS and ShareGPT, for instance, the TVD on `Function' facet falls from 13.05 to 1.57 points, against a noise floor of 1.27.

The two directions are asymmetric.
Matching on classifier scores removes much of the category difference, but matching on categories leaves most of the classifier-detectable dataset signatures intact.
Classifiers therefore capture much of what the taxonomy describes, along with distinctions it does not.
This demonstrates the value of the dataset classification experiment as a measure of separability, with human-designed taxonomies offering a complementary, interpretable view of specific differences.

\vspace{\figneg}
\vspace{\figneg}
\section{Implications For User Modeling}
\label{sec:training}
\vspace{\myneg}

User models are increasingly used to elicit LLM behavior at scale by simulating human--LLM interactions \citep{openai2026gpt56preview}.
To improve simulation fidelity, they are often trained on recorded interactions: \citet{kong2024platolm} train on ShareGPT, \citet{wang2025know} on LMSYS, \citet{naous2025flipping} on WC-1M, and \citet{wang2026learning} on PRISM.
Grounding user models in real data gives them an empirical basis, but leaves open a central question: \textit{how much does the choice of dataset shape the user being simulated?}
Section~\ref{sec:dataset_classification} showed that user messages carry substantial information about their source, including differences that taxonomy-based summaries miss.
A user model trained on such conversations may absorb these dataset-specific regularities, so that models given the same simulation goal produce different interactions depending on their training data.

We examine this possibility through four connected questions.
First, do user models carry their training dataset's signature into the conversations they generate (Section~\ref{sec:training:fingerprint})?
Second, when user models are used to evaluate assistants, does the choice of training dataset change conclusions about the assistants (Section~\ref{sec:training:assistant_evaluation})?
Third, how well do user models trained on one dataset perform on others, and does their ranking depend on the evaluation dataset (Section~\ref{sec:training:cross-dataset})?
Fourth, can dataset classifiers help select training data that improves a user model's fit to a target dataset (Section~\ref{sec:training:guidance})?
Together, these questions cover how user models generate conversations, evaluate assistants, are themselves evaluated, and are trained.

\textbf{User model training.}
Following \citet{naous2025flipping}, we train intent-conditioned user models.
We first annotate each conversation with a high-level intent summarizing the user's goal.
We then fine-tune one model per dataset, starting from the same base model, to generate the next user turn or end the conversation given the intent and the preceding dialogue.
Every user turn serves as a prediction target, with earlier user and assistant turns as context.
This formulation trains the user model to initiate, continue, and terminate synthetic conversations.
We use Qwen2.5-7B-Instruct and Llama-3-8B \citep{grattafiori2024llama} as base models; training and sampling details are in \autoref{appendix:userlm}.

\subsection{Dataset Signatures Persist in Synthetic Conversations}
\label{sec:training:fingerprint}
\vspace{\myneg}

\textbf{Method.}
For each dataset pair, we condition both dataset-specific user models on the same set of conversation intents, drawn from a third dataset used to train neither. Each user model converses with the same assistant (Qwen3.5-9B with reasoning; more results in \autoref{appendix:sec:additional_results}), so the two resulting synthetic conversation sets differ only in the user model. We then apply the dataset classification task in two settings. In the \textit{synthetic-to-synthetic} setting, the classifier is trained on the train split of synthetic conversations. In the \textit{real-to-synthetic} setting, it is trained on the train split of original datasets but tested on the synthetic conversations \citep{mansour2026measuring}. Both settings use the same held-out test split of synthetic conversations and the probe setup of Section~\ref{sec:dataset_classification:observation}, using Qwen2.5-7B embeddings with a trainable logistic regression head.

\textbf{Results}.
As shown in \autoref{tab:fingerprint_propagation},
all ten pairs we tested are distinguishable well above chance in both settings.
Synthetic-to-synthetic accuracy ranges 62.9\% -- 91.5\%,
and real-to-synthetic accuracy 59.3\% -- 75.1\% (chance is 50\%).
The synthetic-to-synthetic results show that user models produce distinguishable conversations even when given identical intents and paired with the same assistant.
Real-to-synthetic accuracy is lower, but remains well above chance: a classifier trained only on real conversations can still tell which dataset a user model was trained on, so the differences reflect characteristics of the original datasets.
Synthetic conversations thus inherit the signatures of their user model's training dataset.
User models trained on different datasets do not simulate interchangeable users; each mirrors the dataset it was trained on.

\begin{table}[H]
\centering
\caption{
    Dataset classification accuracy (\%) on synthetic conversations with Llama-based user models (chance is 50\%).
    For each pair, two user models are conditioned on the same intents from the intent source.
    Real $\to$ real is a reference accuracy, training and testing on real conversations.
    Real $\to$ synth.: train on real, test on synthetic. Synth. $\to$ synth.: train and test on synthetic.
}
\vspace{\figneg}
\vspace{\tablepos}
\label{tab:fingerprint_propagation}
\small
\setlength{\tabcolsep}{6pt}
\renewcommand{\arraystretch}{1.1}
\begin{tabular}{@{}lclcc@{}}
\toprule
Dataset pair
& Real $\to$ real
& Intent source
& Real $\to$ synth.
& Synth. $\to$ synth. \\
\midrule
WC-1M     / WC-4.8M    & 62.8 & HH-RLHF   & 59.3 & 62.9 \\
WC-4.8M   / LMSYS      & 85.1 & WC-1M     & 67.4 & 74.5 \\
WC-4.8M   / ShareChat  & 90.3 & WC-1M     & 73.7 & 79.3 \\
LMSYS     / ShareChat  & 88.7 & WC-4.8M   & 70.0 & 75.8 \\
LMSYS     / ShareGPT   & 83.4 & WC-4.8M   & 63.6 & 72.8 \\
ShareChat / ShareGPT   & 86.8 & LMSYS     & 69.7 & 74.6 \\
ShareChat / Arena      & 84.1 & LMSYS     & 66.7 & 70.3 \\
ShareGPT  / Arena      & 85.6 & ShareChat & 64.9 & 73.3 \\
ShareGPT  / HH-RLHF    & 97.1 & ShareChat & 72.2 & 85.7 \\
Arena     / HH-RLHF    & 96.5 & ShareGPT  & 75.1 & 91.5 \\
\bottomrule
\end{tabular}
\end{table}
\vspace{\figneg}

\subsection{Dataset Signatures Affect The Evaluation of Assistants}
\label{sec:training:assistant_evaluation}
\vspace{\myneg}

User models are increasingly used to evaluate assistants in interactive settings, either on tasks with verifiable answers \citep{laban2026llms} or in open-ended conversations \citep{transluce2026mentalhealth}.
If user models inherit the signature of their training dataset,
the evaluation of an assistant may depend on which dataset trained its simulated users.

\textbf{Tasks with verifiable answers.}
We take 100 math problems from GSM8K \citep{cobbe2021training} and 100 programming problems from HumanEval \citep{chen2021evaluating}.
Each user model receives a problem as its conversation intent and simulates five conversations per problem with each of two assistants,
Qwen2.5-14B-Instruct and Llama-3.1-8B-Instruct.
We apply the user model generation guardrails of \citet{naous2025flipping}: among other constraints, each user turn must contain 3 -- 25 words and must not copy the problem or an earlier turn, so that the original problem is revealed across multiple user turns.
Conversations last at most eight assistant turns.
At the end of simulated conversations, a verifier compares assistant responses with the reference answer (GSM8K) or the unit test (HumanEval).

\begin{table}[H]
\centering
\caption{
    Task success rate (\%) of Qwen2.5-14B-Instruct (Qwen) and Llama-3.1-8B-Instruct (Llama) as assistants when interacting with user models trained on different datasets (rows).
    Each entry averages 500 simulated conversations (100 problems$\times$5).
    90\% CI ($\pm$) is calculated by first averaging the five outcomes for each problem, then using a Student’s t interval across the problem-level averages.
}
\vspace{\figneg}
\vspace{\tablepos}
\label{tab:assistant_task_eval}
\small
\setlength{\tabcolsep}{4pt}
\renewcommand{\arraystretch}{1.1}
\newcommand{\ci}[1]{{\scriptsize$\pm$#1}}
\begin{tabular}{@{}l ccc ccc@{}}
\toprule
& \multicolumn{3}{c}{GSM8K} & \multicolumn{3}{c}{HumanEval} \\
\cmidrule(lr){2-4} \cmidrule(l){5-7}
User model & Qwen & Llama & $\Delta$ & Qwen & Llama & $\Delta$ \\
\midrule
WC-1M     &  9.6\ci{2.6} & 10.0\ci{3.1} & -0.4\ci{2.6} & 14.2\ci{3.2} &  8.4\ci{2.6} &  5.8\ci{3.1} \\
WC-4.8M   & 16.2\ci{4.1} & 13.8\ci{3.4} &  2.4\ci{3.9} & 18.0\ci{3.8} & 11.0\ci{3.2} &  7.0\ci{2.6} \\
LMSYS     & 18.2\ci{4.2} & 17.6\ci{3.5} &  0.6\ci{3.7} & 18.0\ci{4.2} & 14.0\ci{3.7} &  4.0\ci{3.8} \\
ShareChat & 26.8\ci{4.4} & 22.0\ci{4.5} &  4.8\ci{4.5} & 17.0\ci{4.0} & 12.0\ci{3.2} &  5.0\ci{3.1} \\
ShareGPT  & 24.8\ci{3.7} & 14.4\ci{2.9} & 10.4\ci{4.0} & 21.2\ci{4.6} &  9.8\ci{3.1} & 11.4\ci{4.3} \\
Arena     & 13.0\ci{3.4} & 12.4\ci{3.5} &  0.6\ci{2.8} & 14.8\ci{3.5} &  7.2\ci{2.4} &  7.6\ci{3.5} \\
HH-RLHF   & 12.6\ci{3.1} &  9.8\ci{2.5} &  2.8\ci{3.4} & 11.0\ci{2.9} &  7.8\ci{2.4} &  3.2\ci{2.7} \\
\bottomrule
\end{tabular}
\end{table}
\vspace{\figneg}

\textbf{Results.}
\autoref{tab:assistant_task_eval} shows that measured task success depends on the dataset used to train the user model.
Even with the problems fixed, each assistant's success varies by up to 2.8$\times$ across user models: from 9.6\% to 26.8\% for Qwen2.5-14B-Instruct on GSM8K, and from 7.2\% to 14.0\% for Llama-3.1-8B-Instruct on HumanEval.
We attribute this variation to differences in how user models simulate conversations, such as how models approximate users in intent expression and terminology use \citep{piantadosi2012communicative}, which are inherited from the training data.

The user model also changes the overall takeaway:
whether the larger assistant (Qwen2.5-14B-Instruct) appears meaningfully stronger than the smaller one (Llama-3.1-8B-Instruct) or if their gap is trivial \citep{chang-etal-2025-chatbench}.
On GSM8K, the two perform almost identically with the WC-1M user model (9.6\% vs.\ 10.0\%).
With the ShareGPT user model, however, Qwen2.5-14B-Instruct outperforms Llama-3.1-8B-Instruct (24.8\% vs.\ 14.4\%).
Across both benchmarks, the ShareGPT user model separates the two assistants the most (10.4 and 11.4 points), whereas the LMSYS user model yields some of the smallest margins (0.6 and 4.0 points).
Which assistant appears better, and by how much, depends on the dataset used to train the user model.

\subsection{Cross-dataset Generalization}
\label{sec:training:cross-dataset}
\vspace{\myneg}

The persistence of dataset signatures in synthetic conversations raises a further question: how well do user models generalize beyond their training source?
In computer vision, \citet{liu2025decade} and \citet{tommasi2015deeperlookdatasetbias} have shown that strong within-dataset performance need not translate to other datasets.
We test this for user models by evaluating each dataset-specific user model on the test split of every dataset. We use the perplexity of user turns as a standard measure of user model quality \citep{shuster2021dialogue,naous2025flipping,moon2026identity}.

\begin{table}[H]
\centering
\caption{Cross-dataset evaluation. Rows indicate training and columns indicate test (target) datasets. Each entry represents mean per-token perplexity measured on the test split. Bold and underlined entries indicate the lowest and second-lowest perplexity in each column, respectively.}
\label{tab:cross_dataset_evaluation_2}
\vspace{\figneg}
\vspace{\tablepos}
\small
\setlength{\tabcolsep}{4pt}
\renewcommand{\arraystretch}{1.15}
\resizebox{0.85\textwidth}{!}{
\begin{tabular}{@{}l*{7}{r}@{}}
\toprule
    \shortstack[l]{Train $\downarrow$ Test $\rightarrow$}
    & WC-1M
    & WC-4.8M
    & LMSYS
    & Arena
    & ShareChat
    & ShareGPT
    & HH-RLHF \\
    \midrule
    WC-1M
    & \textbf{6.41} & \underline{6.70} & \underline{6.14} & 6.15 & 8.38
    & \underline{6.02} & 5.39 \\

    WC-4.8M
    & \underline{6.53} & \textbf{6.36} & 6.17 & \underline{6.06} & \underline{8.28}
    & 6.07 & 5.46 \\

    LMSYS
    & 6.61 & 6.80 & \textbf{5.43} & 6.17 & 8.31
    & 6.03 & \underline{5.19} \\

    Arena
    & 6.69 & 6.82 & 6.22 & \textbf{5.75} & 8.29
    & 6.14 & 5.29 \\

    ShareChat
    & 6.71 & 6.85 & 6.22 & 6.13 & \textbf{7.33}
    & 6.09 & 5.30 \\

    ShareGPT
    & 6.65 & 6.92 & 6.21 & 6.26 & 8.40
    & \textbf{5.62} & 5.28 \\

    HH-RLHF
    & 7.05 & 7.38 & 6.72 & 6.80 & 9.07
    & 6.48 & \textbf{4.53} \\
    \bottomrule
\end{tabular}
}
\end{table}
\vspace{\figneg}

\textbf{Results.}
\autoref{tab:cross_dataset_evaluation_2} reports results for user models fine-tuned from Qwen2.5-7B-Instruct (Llama-3-8B results in \autoref{tab:cross_dataset_evaluation}).
On every test dataset, the user model trained on that dataset achieves the lowest perplexity, producing a diagonal pattern.
This in-domain advantage shows that performance depends strongly on the similarity between training and evaluation sources. 
The ranking of user models also varies with the evaluation dataset:
not only the top-performing model (the diagonal), but also the second-best model;
for example, the user model trained with WC-4.8M is ranked second on WC-1M, Arena, and ShareChat, but it is the worst-performing model on HH-RLHF. 
Thus, our understanding of user model quality depends on what dataset it is tested on.

HH-RLHF provides an interesting example of the distinction between in-domain fit and cross-dataset transfer. The HH-RLHF user model achieves an in-domain perplexity of only 4.53 (well below other in-domain perplexities of 5.43 -- 7.33), yet has the worst performance on every other dataset. Its low in-domain perplexity may indicate dataset-specific regularities, such as formulaic crowdworker prompts.
Differences in in-domain perplexity should be interpreted cautiously:
they may reflect how predictable each dataset's conversations are, rather than how well each model simulates users.

\textbf{Separability correlates with the transfer gap.}
We next ask whether the separability of two datasets predicts how poorly user models transfer between them. For each of the 21 dataset pairs $(A,B)$, we train a binary linear probe on Qwen2.5-7B representations, recording its accuracy on the test split (chance is 50\%).
We compare this accuracy with the symmetric transfer gap $\Delta_{AB}$ in \autoref{eq:sym_transfer_gap}
where $\mathrm{NLL}({A\to B})$ is the mean per-token negative log-likelihood (i.e., log perplexity) that a user model trained on $A$ assigns to the user turns of $B$. $\Delta_{AB}$ is the excess loss incurred by using the out-of-domain user model in place of the in-domain one, averaged over both directions.

\begin{equation}
\centering
\label{eq:sym_transfer_gap}
\Delta_{AB}
= \frac{1}{2}\Big[
    \big(\mathrm{NLL}(A \to B) - \mathrm{NLL}(B \to B)\big)
    +
    \big(\mathrm{NLL}(B \to A) - \mathrm{NLL}(A \to A)\big)
\Big].
\end{equation}

\setlength{\intextsep}{0pt}
\begin{wrapfigure}{r}{0.5\textwidth}
    \centering
    \includegraphics[width=1.00\linewidth]{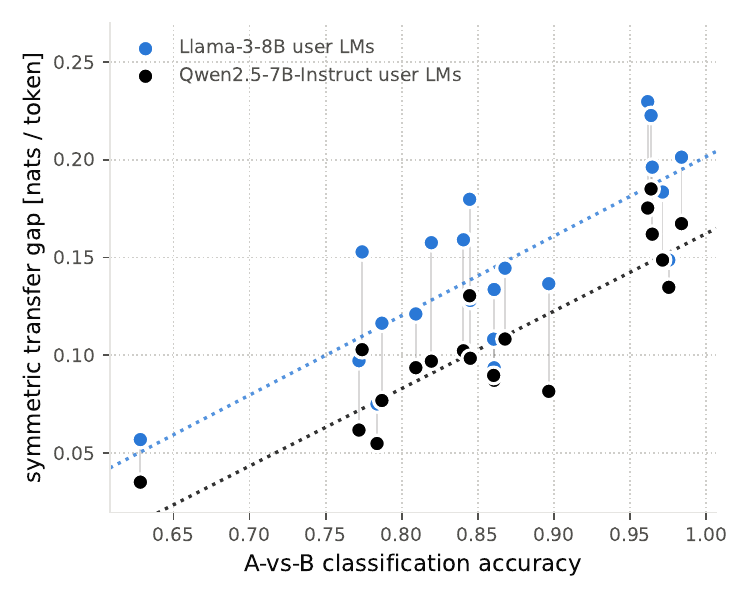}
    \vspace{\figneg}
    \vspace{\figneg}
    \caption{
    Symmetric transfer gap $\Delta_{AB}$ correlates with the $A$-to-$B$ dataset classification accuracy.
    }
    \label{fig:sep_vs_gap}
\end{wrapfigure}

Figure~\ref{fig:sep_vs_gap} shows a strong positive correlation: Pearson $r = 0.80$ (Spearman $\rho = 0.72$) for Llama-based user models and $r = 0.86$ ($\rho = 0.81$) for Qwen-based user models ($p < 10^{-3}$ in all cases).
The least separable pair, WC-1M vs.\ WC-4.8M (accuracy 63\%), has the smallest gap, whereas pairs involving HH-RLHF (accuracy $\geq$ 95\%) have among the largest. Qwen-based user models have consistently smaller transfer gaps than Llama-based ones but preserve a similar ordering across pairs.
The strong correlation demonstrates a practical advantage of dataset classification:
using a linear probe on language representations, we can get a cheap proxy for how well a user model trained on one dataset will transfer to the other, before any training.

\subsection{Classifier-guided Data Selection}
\label{sec:training:guidance}
\vspace{\myneg}

Section~\ref{sec:training:cross-dataset} showed that the choice of training conversations affects a user model's fit to a target dataset, and that separability correlates with this effect. Can dataset classifiers, then, help select useful training data for a given target dataset under a fixed training budget?

\textbf{Method.}
We adapt data selection via importance resampling (DSIR; \citet{xie2023data}).
For each \textit{target} dataset, we pool the training splits of the other six datasets into a \textit{donor} pool.
We train a binary classifier to distinguish target conversations from donor conversations.
Let $s_T(c)$ be classifier-predicted probability that a donor conversation $c$ belongs to the target. We sample donor conversations without replacement, with probability proportional to

\vspace{\myneg}
\begin{equation}
    w_T(c)
    = \frac{s_T(c)}{1-s_T(c)}
    = \frac{\text{Prob(classifies as target)}}{\text{Prob(classifies as donor)}},
    c \in \text{donor pool}
    \label{eq:donor_selection_weight}
\end{equation}
\vspace{\myneg}

\begin{figure}[H]
    \centering
    \includegraphics[width=1.00\linewidth]{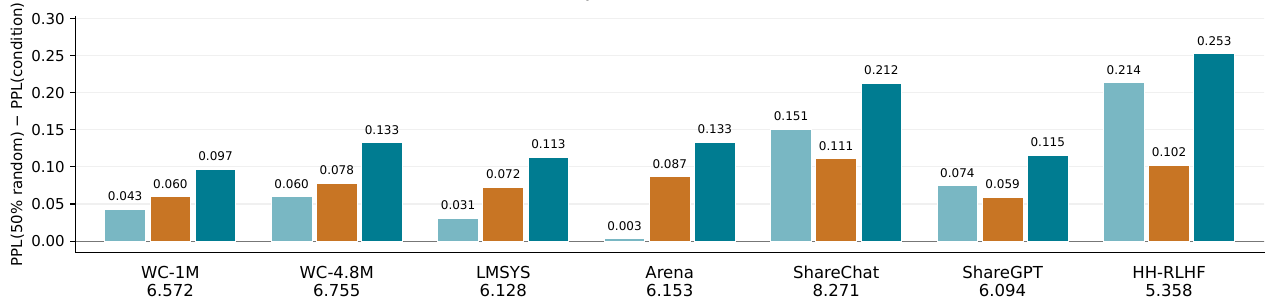}
    \vspace{\figneg}
    \vspace{\figneg}
    \caption{
    Perplexity reduction relative to uniform random sampling at 50\% of the training budget,
    PPL(50\% random) - PPL(condition), for Qwen2.5-7B-Instruct user models (higher is better).
    The number under each dataset is the 50\%-random perplexity.
    Classifier-guided selection outperforms random sampling at every matched budget. Llama-3-8B results are in \autoref{fig:classifier-guided-ppl-llama}.
    }
    \label{fig:classifier-guided-ppl}
\end{figure}

We compare this classifier-guided data selection (labeled `DSIR' in \autoref{fig:classifier-guided-ppl}) with uniform random sampling from the donor pool, each at $50\%$ and $100\%$ of the training sample budget (40K conversation samples). Both conditions train the user model exclusively on donor conversations.

\vspace{10pt}

\textbf{Results.}
At matched training budgets, classifier-guided data selection yields lower target perplexity than random sampling (Figure~\ref{fig:classifier-guided-ppl}).
For three of the seven targets, selection at 50\% budget also beats random sampling at 100\%.
For HH-RLHF, for example, it reduces perplexity from 5.36 to 5.14 at the 50\% budget, compared with 5.26 for random sampling with twice the data.
These results show that the classifier identifies donor conversations that better support fitting user utterances in the target.
\section{Related Work}
\label{sec:prelim}
\vspace{\myneg}

\subsection{Dataset Signatures}
\label{sec:prelim:dataset_bias}
\vspace{\myneg}

Dataset classification experiments and cross-dataset evaluations have exposed source-specific regularities in vision datasets
\citep{torralba2011unbiased, tommasi2015deeperlookdatasetbias},
including modern large-scale collections \citep{liu2025decade}.
Such regularities make within-dataset performance an unreliable guide to generalization.
NLP has analogous findings:
annotation protocols introduce predictive shortcuts \citep{gururangan2018annotation},
and corpus filtering changes which content and populations are disproportionately represented \citep{dodge2021documenting}.
\citet{mansour2026measuring} identify fingerprints of web-filtered corpora
and show that they propagate through LLM pretraining.
However, human–LLM dataset releases often emphasize diverse real-world interactions \citep{zhao2024wildchat, zheng2024lmsys} yet dataset separability in this setting is less understood.
We address this gap by examining three aspects:
dataset signatures beyond human-designed taxonomy categories \citep{shelby2025taxonomy},
their persistence through user model training,
and their relationship to cross-dataset generalization.

\subsection{Human-LLM Interaction Datasets}
\label{sec:prelim:human_llm}
\vspace{\myneg}

Human--LLM datasets support both descriptive research and model development. They inform studies of AI use, such as Anthropic Economic Index \citep{handa2025economic}, and taxonomies of user behavior \citep{shelby2025taxonomy, shaikh2025navigating}. For example, AI Observatory examines how descriptions of real-world AI use vary across conversation sources \citep{longpre2026ai}.
For assistant development, recorded queries and model responses provide supervision for assistant fine-tuning, as in Vicuna \citep{vicuna2023}. Additionally, preference alignment uses conversation logs paired with preference labels, motivating HH-RLHF's collection \citep{bai2022training}. For evaluation, WildBench constructs a real-world task benchmark from the WildChat collection \citep{lin2025wildbench}.

Recently, user models (also called user simulators) have gained attention \citep{humansand2026persimmon} as a scalable way to train and evaluate assistants through adaptive multi-turn interactions \citep{kong2024platolm, naous2025flipping, wang2026learning}. They support evaluations of assistant behaviors emerging over extended conversations \citep{openai2026gpt56preview} and assistant training through simulated interactions \citep{suh2026quantifying}. We therefore focus the second half of this paper on user modeling: it directly transforms recorded user behavior into simulated interactions used for downstream experiments. This opens up the possibility that the choice of training dataset is consequential for both the simulated user behavior and the conclusions about assistant behaviors drawn from simulation.
\vspace{\myneg}
\section{Conclusion, Limitations, and Future Work}
\label{sec:conclusion}
\vspace{\myneg}

By revisiting the dataset classification experiment of \citet{torralba2011unbiased},
we show that human-LLM interaction datasets contain signatures that survive
marker removal, assistant masking, and matching on human-designed taxonomy summaries.
Then, we investigate the implications of dataset signatures for user modeling.
Dataset signatures persist in conversations generated by user models;
the evaluation of interactive assistants and user models depends on which dataset we use to train user models.
Dataset classifiers also have a practical use: classifier-guided selection improves a user model's fit to a target dataset.
Together, these findings demonstrate that dataset choice is consequential both for the user behavior being simulated and for the evidence used to assess it.
Our results motivate documenting collection context and testing research findings across datasets.

Our study has several limitations. We study English conversations from public datasets, so application of our findings to multilingual, cross-cultural domains is an open problem.
We also note that dataset separability neither identifies one `representative' dataset nor isolates the causes of source differences:
recruitment, incentives, collection periods, and more can vary together.
Matching selected taxonomy summaries does not control for all semantic or joint distributional differences.
Perplexity measures predictive fit, rather than human judgments of simulation fidelity which should be verified by a separate user study process.
Future work could disentangle collection effects, extend evaluation across languages and populations, and test whether better predictive fit yields more realistic simulators or reliable assistant evaluations.

\subsubsection*{Acknowledgments}

We are grateful to Prof. John Canny, David Chan, and Zhining Zhang for their thoughtful comments and valuable feedback.
Joseph Suh acknowledges support from the Korea Foundation for Advanced Studies through the Doctoral Study Abroad Program.
This work was supported by compute resources from the Center for Human-Compatible AI (CHAI).
The views and findings expressed herein are those of the authors and do not reflect the positions of any sponsoring organization.

\bibliography{iclr2027_conference}
\bibliographystyle{iclr2027_conference}

\appendix

\textbf{Code release.} Our code is avaialble at the following link: \\
\url{https://github.com/JosephJeesungSuh/dataset-signature-human-llm}

\section{Dataset Curation}
\label{appendix:dataset}

\subsection{Datasets}
\label{appendix:dataset:explanation}

Datasets are ordered chronologically.

\textbf{HH-RLHF} \citep{bai2022training} is a human preference dataset collected by Anthropic through crowdworkers' interactions with AI assistants under two task conditions: (1) helpfulness and (2) harmlessness.
(1) In the helpfulness condition, participants engaged in open-ended conversations, such as question answering, writing, and planning. At each assistant turn, participants compared two candidate responses and selected the response that is considered more helpful.
(2) In the harmlessness condition, participants were instructed to adversarially probe the models to elicit harmful responses and select the more harmful candidate. Conversations continued from the selected response, allowing participants to produce multi-turn interactions.

\textbf{ShareGPT} contains user-shared ChatGPT conversations collected from ShareGPT.com. ShareGPT conversations were subsequently used to train Vicuna \citep{zheng2023judging}. Data filtering included converting HTML content to Markdown, filtering inappropriate or low-quality examples, and splitting lengthy conversations into segments compatible with the model’s context window. Because multiple snapshots and processed derivatives circulate under the ShareGPT name on HuggingFace, we identify our input as the two HTML-cleaned JSON files in anon8231489123/ShareGPT\_Vicuna\_unfiltered, containing 77K conversation records before our curation.

\textbf{LMSYS-Chat-1M} \citep{zheng2024lmsys} was collected from voluntary user interactions on the Vicuna demonstration and Chatbot Arena website between April and August 2023. The platform provided free access through three interfaces: a single-model chat, a comparison between two randomly selected anonymous models, and a comparison between two user-selected models. The dataset comprises one million conversations with 25 models, originating from 210K unique users. Before release, the researchers removed conversations containing personally identifiable information (PII) and added message-level moderation annotations. The released corpus contains conversation logs, without the human preference labels available in a separate Chatbot Arena dataset.

\textbf{Arena Human Preference 140K} extends the collection setting of LMSYS-Chat-1M by pairing user–model interactions with explicit human preference judgments. Whereas LMSYS-Chat-1M contains conversations collected across single-model and comparison interfaces in 2023, Arena contains model comparisons collected between April and July 2025. Users submitted their own prompts and voted for either anonymous model, a tie, or dissatisfaction with both. The updated interface allowed users to continue a conversation after voting, with a newly sampled model pair receiving prior conversational context. Each record therefore represents a preference judgment within a potentially longer session, rather than an independent conversation. The release contains 135,634 vote records with paired transcripts, model identifiers, and session metadata.

\textbf{WildChat-1M} \citep{zhao2024wildchat} was collected through publicly accessible chatbot services hosted on HuggingFace Spaces and powered by GPT-3.5-Turbo and GPT-4. Users received free access after explicitly consenting to the collection and sharing of their interactions. Conversations are based on users’ requests, not on researcher-assigned topics. The collection spans April 2023 to May 2024 and includes both single-turn and multi-turn interactions. Individual conversation records were linked into conversations using conversation histories, IP addresses, and request headers. Before release, the researchers applied automated procedures to redact PII, hashed IP addresses, and derived coarse geographic metadata. After reserving a subset for benchmark development, the release contained approximately 1 million conversations.

\textbf{WildChat-4.8M} \citep{zhao2024wildchat} extends the WildChat-1M collection through continued logging of voluntary interactions with publicly accessible chatbot services, with coverage extending through July 2025. The expanded collection incorporates additional OpenAI model families, including GPT-4o, GPT-4.1-mini, and the o1 reasoning models. Release preparation included automated privacy redaction, scanning for exposed secrets, and content moderation. The collection originally comprised approximately 4.8 million conversations; the public release named WildChat-4.8M applies toxicity filtering and contains approximately 3.2 million conversations.

\textbf{ShareChat} \citep{yan2025sharechat} was constructed by retrospectively collecting publicly shared conversations from five commercial chatbot platforms: ChatGPT, Claude, Gemini, Perplexity, and Grok. The researchers identified shared conversation URLs through the Wayback Machine using platform-specific URL patterns, then employed automated browser interactions to retrieve rendered pages and extract structured conversation records. Users had independently conducted and shared these conversations, so the collection did not involve researcher-assigned conversational tasks. Extraction preserved ordered user–assistant exchanges and available platform-specific features, including citations, visible thinking traces, code artifacts, timestamps, and model identifiers. Personally identifiable information was redacted using automated tools, with an additional model-based audit of the redaction results. The reported corpus contains 143K conversations and 660K interaction turns, spanning 95 languages and usage between April 2023 and October 2025.

\subsection{Curation and split construction}
\label{appendix:dataset:curation}

We convert each source to alternating user--assistant messages.
HH-RLHF uses the chosen transcript according to the preference label.
For Arena, we retain one longest released history per evaluation session;
the stored assistant text is the displayed left-side view, rather than a reconstructed sequence of the responses selected by the user.
ShareChat records with ambiguous message ordering are excluded.

We use released language labels when available and fastText \citep{joulin2017bag} language identification otherwise.
Table~\ref{tab:dataset_attrition} reports the size of input dataset snapshots and retained sample sizes.

\begin{table}[t]
\centering
\small
\setlength{\tabcolsep}{3pt}
\caption{Curation and sampling counts for the snapshots used in this study. Input counts are raw records. English counts follow conversation reconstruction, structural validation, and language filtering. Eligible counts additionally reflect duplicate/template filtering and cross-source exclusions.}
\resizebox{0.6\columnwidth}{!}{%
\begin{tabular}{@{}lrrrr@{}}
\toprule
Dataset & Input & English & Eligible & Retained \\
\midrule
HH-RLHF & 169,352 & 168,805 & 145,740 & 50,000 \\
ShareGPT & 76,920 & 54,352 & 50,790 & 50,000 \\
LMSYS-Chat-1M & 1,000,000 & 766,545 & 415,087 & 50,000 \\
Arena 140K (2025) & 135,634 & 60,810 & 57,184 & 50,000 \\
WildChat-1M & 837,989 & 473,246 & 285,771 & 50,000 \\
WildChat-4.8M & 3,199,860 & 1,672,053 & 225,212 & 50,000 \\
ShareChat & 129,584 & 79,453 & 75,254 & 50,000 \\
\bottomrule
\end{tabular}}
\label{tab:dataset_attrition}
\end{table}

\textbf{Duplicate and template filtering.}
For duplicate matching, we lowercase user text and collapse whitespace.
We remove exact duplicates of concatenated user text within each source, and filter opening prompt templates using 7-grams occurring more than 100 times within a source.
WC-4.8M excludes conversation hashes present in WC-1M.
Full user texts of at least five words appearing in multiple sources are excluded as well.
\section{Neural Classifier}
\label{appendix:classifier}
\vspace{\myneg}

All dataset classifiers share one recipe: a linear probe on frozen representations of a language model.

\textbf{Input to the language model.}
We format each conversation as alternating lines of the form \texttt{User: ...} and \texttt{Assistant: ...}, separated by newlines.
Dataset-specific markers (\autoref{tab:datasets_main}) are removed beforehand.
Unless stated otherwise, every assistant response is replaced with the placeholder \texttt{<assistant\_response>}, so the classifier sees only user messages and the turn structure.

\textbf{Representations.}
We encode each conversation with the frozen Qwen2.5-7B and read its hidden states at layers 6, 12, 18, and 28 (the final layer).
From each layer, we form two representations: the mean of the hidden states over all tokens, and the hidden state at the final EOS token.
This yields eight candidate representations per conversation.
For the model size comparison (\autoref{fig:observation_scaling}),
we apply the same procedure to other Qwen2.5 family models,
using four layers that span the network's depth and include the final layer
(e.g., layers 12, 24, 36, and 48 for Qwen2.5-14B which has 48 layers).

\textbf{Classifier head.}
On each representation, we train a multinomial logistic regression head.
We standardize features with the mean and standard deviation of the training split.
The loss is cross-entropy with class weights inversely proportional to class frequency,
plus an L2 penalty on the weight matrix.
We optimize it with Adam (learning rate 0.05) for 300 epochs from a zero initialization,
computing validation balanced accuracy every 10 epochs and keeping the best checkpoint.

\textbf{Model selection and evaluation.}
For each classification task, we select the representation (layer and pooling), the regularization strength $\lambda \in \{10^{-4}, 10^{-3}, 10^{-2}\}$, and the checkpoint by validation accuracy.
Each dataset contributes 40K training, 3K validation, and 7K test conversations,
so test sets are balanced and accuracy equals balanced accuracy.
In the real-to-synthetic setting, training and model selection use only real conversations,
and synthetic conversations are used only for testing.

\section{Taxonomization and classifier-score matching}
\label{app:taxonomization}
\label{appendix:taxonomization}

\subsection{Taxonomy annotations}

\begin{figure}[t]
\centering
\includegraphics[width=1.0\linewidth]{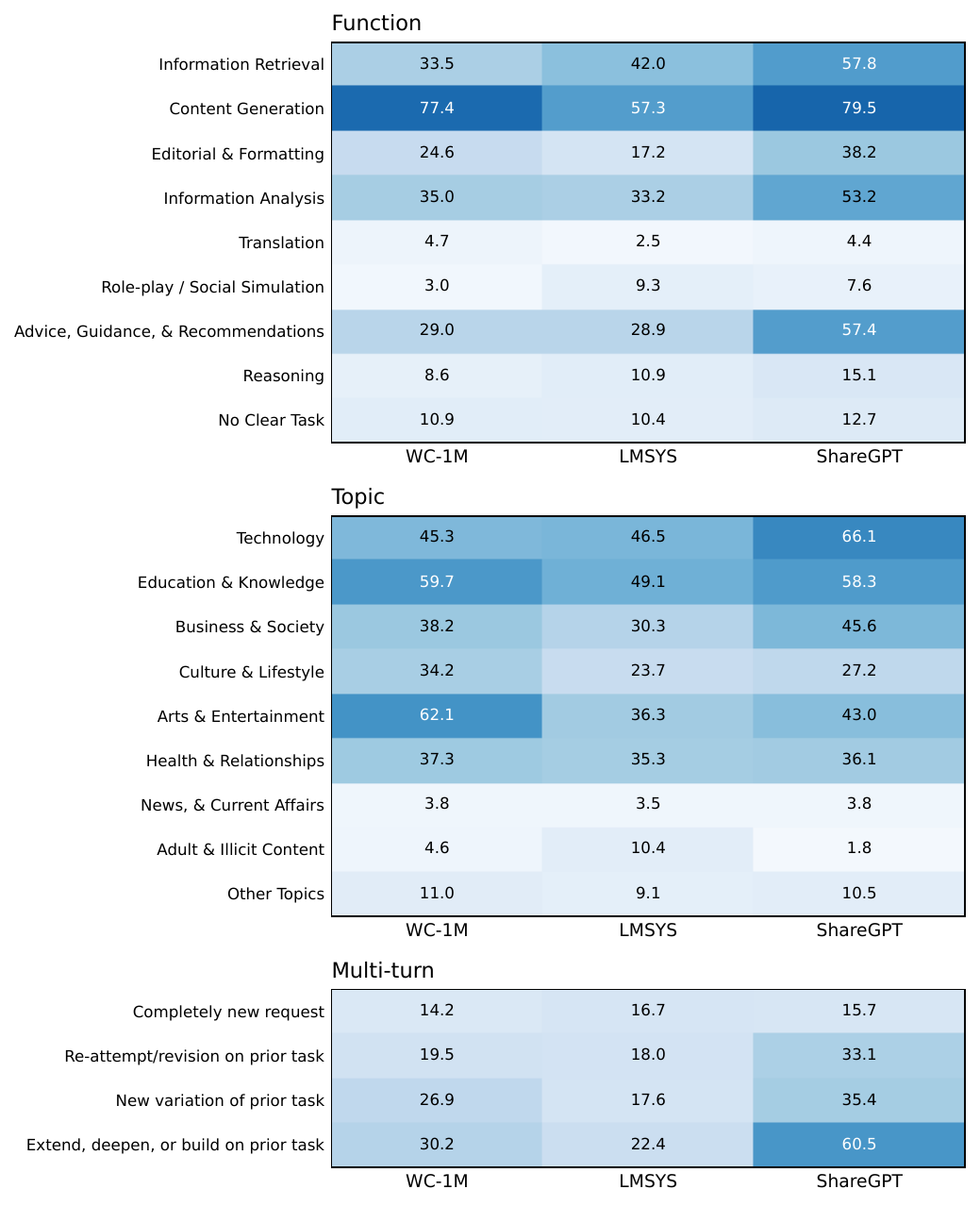}
\caption{Percentage of conversations containing each annotated category in the original held-out WC-1M, LMSYS, and ShareGPT cohorts. A conversation can contain several categories, so columns need not sum to 100\%. Each factor uses only conversations with complete accepted annotations.}
\vspace{\figneg}
\vspace{\figneg}
\label{fig:taxonomy_prevalences}
\end{figure}

The three factors describe what a user asks the assistant to do (Function), what the request is about (Topic), and how a follow-up relates to the preceding task (Multi-turn), using the AI Observatory taxonomy \citep{longpre2026ai}. Function and Topic each aggregate to nine broad categories. The four follow-up relationships are extending/deepening the prior task, retrying/revising it, requesting a variant, and making a completely new request. Figure~\ref{fig:taxonomy_prevalences} shows the measured category prevalences.

\textbf{Annotations.}
We apply the AI Observatory taxonomy using gpt-5.6-luna with a confidence threshold of 0.7.
The annotation procedure is based on their official codebase without any modification.

\subsection{Matching attributes}
\vspace{\myneg}

\textbf{Maximum-size attribute matching.}
Let $x_i,y_j\in\{0,1\}$ indicate selection from sources $A$ and $B$, and let $f_k$ be a selected attribute.
We solve

\begin{equation}
\begin{aligned}
\max_{x,y}\;\sum_i x_i \quad \text{subject to }&\sum_i x_i=\sum_j y_j,\\
&\sum_i f_k(i)x_i=\sum_j f_k(j)y_j\quad\text{for every selected }k.
\end{aligned}
\end{equation}
\vspace{\figneg}

For the structural factor, the features are user-turn count and total user-text token count.
Equal conversation counts and equal feature totals imply equal mean user turns per conversation and equal total user tokens divided by total user messages. User-text tokens are counted with cl100k\_base.

\subsection{Within-dataset split-half bootstrapping}
\label{appendix:taxonomization:split_half}
\vspace{\myneg}

To quantify the noise floor of taxonomy category distributions,
we repeatedly divide each original test dataset and each pair-specific score-matched dataset into two disjoint halves. Each of 1,000 repetitions draws a random permutation and takes two groups of $\lfloor N/2\rfloor$ conversations. We repeatedly compute the distributional difference between two halved subsets and calculate the median of bootstrap results to obtain the estimate of noise floor.
\vspace{\figneg}
\section{User Model Training}
\label{appendix:userlm}
\vspace{\figneg}

\textbf{Intent Generation.}
For each conversation, we generate a high-level user intent summary, following \citet{naous2025flipping}. Specifically, we provide the full human--LLM conversation trajectory to Qwen3-32B and instruct it to produce a short summary of the user's high-level intent. We refer the reader to the ``Prompt Template for Intent Generation'' in \citet{naous2025flipping} for the exact prompt.

\textbf{Supervised Fine-Tuning.}
We perform supervised fine-tuning (SFT) on the user-side utterances of the training split.
Below, we present a training sample from WC-1M formatted with the Qwen-2.5 chat template,
where highlighted spans indicate the tokens over which the loss is computed:

{
\small
\noindent
\begin{tabular}{@{}p{\textwidth}@{}}
\texttt{<|im\_start|>system} \\
\texttt{You are a user chatting with an assistant language model to convert a website into a PDF format.<|im\_end|>} \\
\texttt{<|im\_start|>user} \\
\hl{\texttt{To make the following website as an PDF<|im\_end|>}} \\
\texttt{<|im\_start|>assistant} \\
\texttt{To convert a website to a PDF, you can follow these steps: \ldots (truncated) \ldots but not its dynamic or interactive elements.<|im\_end|>} \\
\texttt{<|im\_start|>user} \\
\hl{\texttt{Can you do that for me?<|im\_end|>}} \\
\texttt{<|im\_start|>assistant} \\
\texttt{Apologies for the confusion, but as a text-based AI, \ldots (truncated) \ldots to assist you.<|im\_end|>} \\
\texttt{<|im\_start|>user} \\
\hl{\texttt{<|endconversation|><|im\_end|>}} \\
\texttt{<|endoftext|>}
\end{tabular}
}

We explain some details.
First, the prepared intent (``You are a user chatting\ldots'') is placed in the system prompt to condition generation.
Second, the loss is applied not only to user-turn content but also to the turn-ending special token (\texttt{<|im\_end|>} for Qwen2.5-family models), so that the model learns to produce delimited turns.
Third, following \citet{naous2025flipping}, we introduce a special token \texttt{<|endconversation|>} to signal that the user has terminated the conversation:
a real human--LLM conversation with $2n$ turns is structured into $2n+1$ turns with $n+1$ user-side utterances (the last one always being \texttt{<|endconversation|>}) and $n$ LLM responses.
We initialize the embedding of the special token with the pretrained EOS token embedding.

\textbf{Synthetic Conversation Generation.}
The fingerprint experiments use a fixed Qwen3.5-9B assistant with thinking disabled. Only the user model sees the intent; the assistant receives the generated dialogue without an additional system prompt. Each intent produces one conversation per user model. The initial user request is generated, rather than copied from the seed conversation. Train, validation, and test splitting follows the seed intent's original split, and the two synthetic conversation datasets share the same intents within each split.

User sampling uses temperature 0.7, top-$p$ 0.9, and no top-$k$ restriction.
Assistant sampling uses temperature 0.7, top-$p$ 0.8, top-$k$ 20.
The served user-model context is 8,192 tokens.
A conversation stops at the user termination token, four completed assistant replies, or the context limit.
A user utterance identical to the immediately preceding user utterance is resampled up to seven times, increasing temperature by 0.05 per retry to a maximum of 1.0.
\section{Additional Results}
\label{appendix:sec:additional_results}

\subsection{More ablations}

\vspace{6pt}
\begin{table}[H]
\centering\small
\caption{Frozen language models vs.\ co-trained classifiers on WC-1M + LMSYS + ShareChat.}
\label{tab:observation_finetuning}
\vspace{\figneg}
\begin{tabular}{lrrr}
\toprule
Model size & Frozen & Fine-tuned & $\Delta$ [\%] \\
\midrule
0.5B & 69.7 & 80.3 & +10.6 \\
1.5B & 70.1 & 81.4 & +11.3 \\
3B   & 72.5 & 80.2 & +7.7  \\
7B   & 72.5 & 79.3 & +6.8  \\
\bottomrule
\end{tabular}
\end{table}

\textbf{Ablations: Co-training language models with classifier heads.}
Co-trained models outperforms their corresponding frozen language model configurations by 6.8--11.3 percentage points (Table~\ref{tab:observation_finetuning}).


\subsection{Dataset classification on Synthetic Conversations: More Assistants}
\label{appendix:sec:fingerprint_assistants}

Section~\ref{sec:training:fingerprint} pairs every user model with Qwen3.5-9B as the assistant.
To test whether the results depend on this assistant choice, we repeat the experiment with Llama-3.1-8B-Instruct and Qwen2.5-14B-Instruct as assistants, while fixing the set of intent sources and dataset pairs.

\autoref{tab:fingerprint_assistants} shows that accuracy barely changes with the assistant:
Averaged over the ten pairs, synthetic-to-synthetic accuracy is 76.1\%, 75.4\%, and 75.7\% with Qwen3.5-9B, Llama-3.1-8B-Instruct, and Qwen2.5-14B-Instruct, respectively, and real-to-synthetic accuracy is 68.3\%, 68.5\%, and 68.3\%.
This indicates that the signature of a user model's training dataset persists regardless of which assistant it interacts with.

\vspace{10pt}
\begin{table}[H]
\centering
\caption{
    User-model classification accuracy (\%) on synthetic conversations with three different assistants (chance is 50\%).
    For each pair, two user models are conditioned on the same intents from the intent source and interact with the same assistant;
    the Qwen3.5-9B columns repeat \autoref{tab:fingerprint_propagation}.
    Real $\to$ synth.: train on real, test on synthetic. Synth. $\to$ synth.: train and test on synthetic.
}
\vspace{\figneg}
\vspace{\tablepos}
\label{tab:fingerprint_assistants}
\small
\setlength{\tabcolsep}{4pt}
\renewcommand{\arraystretch}{1.1}
\begin{tabular}{@{}ll ccc ccc@{}}
\toprule
& & \multicolumn{3}{c}{Real $\to$ synth.} & \multicolumn{3}{c}{Synth. $\to$ synth.} \\
\cmidrule(lr){3-5} \cmidrule(l){6-8}
Dataset pair & Intent source & Qwen3.5 & Llama3.1 & Qwen2.5 & Qwen3.5 & Llama3.1 & Qwen2.5 \\
\midrule
WC-1M / WC-4.8M       & HH-RLHF   & 59.3 & 58.8 & 59.2 & 62.9 & 62.6 & 64.2 \\
WC-4.8M / LMSYS       & WC-1M     & 67.4 & 67.5 & 66.9 & 74.5 & 72.9 & 74.3 \\
WC-4.8M / ShareChat   & WC-1M     & 73.7 & 73.9 & 74.0 & 79.3 & 78.8 & 78.4 \\
LMSYS / ShareChat     & WC-4.8M   & 70.0 & 70.6 & 70.3 & 75.8 & 76.4 & 77.3 \\
LMSYS / ShareGPT      & WC-4.8M   & 63.6 & 64.8 & 65.2 & 72.8 & 70.6 & 70.0 \\
ShareChat / ShareGPT  & LMSYS     & 69.7 & 69.6 & 69.2 & 74.6 & 73.6 & 73.6 \\
ShareChat / Arena     & LMSYS     & 66.7 & 67.3 & 66.8 & 70.3 & 70.9 & 71.5 \\
ShareGPT / Arena      & ShareChat & 64.9 & 64.6 & 63.8 & 73.3 & 71.2 & 71.0 \\
ShareGPT / HH-RLHF    & ShareChat & 72.2 & 72.3 & 72.2 & 85.7 & 85.1 & 85.3 \\
Arena / HH-RLHF       & ShareGPT  & 75.1 & 75.3 & 75.3 & 91.5 & 91.6 & 91.7 \\
\bottomrule
\end{tabular}
\end{table}

\subsection{Dataset classification on Synthetic Conversations: Between Assistants}
\label{appendix:sec:assistant_identity}

Section~\ref{sec:training:fingerprint} varies the user model while fixing the assistant;
the previous section shows additional results with multiple assistants.
Here we \textit{fix the intents and the user model and vary the assistant}.

For each of 17 combinations of intent source and user model from the setup of Section~\ref{sec:training:fingerprint}, the same user model converses with Qwen3.5-9B, Qwen2.5-14B-Instruct, and Llama-3.1-8B-Instruct on the same intents.
We train classifiers to identify the assistant in three binary comparisons and one three-way comparison.
As in the main experiments, assistant responses are masked, so the classifier must identify the assistant only through its influence on the user model's multi-turn messages.
We evaluate on all conversations, and separately on single-turn conversations (one user message) and multi-turn conversations (two or more user messages).
Because these subsets differ in size across assistants, we report balanced accuracy.
To measure how distinguishable the assistants themselves are, we also train classifiers on assistant responses alone, masking the user turns.

\begin{table}[H]
\centering
\caption{
    Assistant classification accuracy (\%) on synthetic conversations with the same intents and user model.
    Entries are means over 17 combinations of intent source and user model, with the range in parentheses.
    The first three columns use user turns only, with assistant responses masked; single-turn conversations have one user message and multi-turn conversations two or more.
    The last column uses assistant responses only, with user turns masked.
}
\vspace{\figneg}
\vspace{\tablepos}
\label{tab:assistant_identity}
\small
\setlength{\tabcolsep}{4pt}
\renewcommand{\arraystretch}{1.15}
\begin{tabular}{@{}lccccc@{}}
\toprule
& & \multicolumn{3}{c}{User turns only} & Assistant \\
\cmidrule(lr){3-5}
Comparison & Chance & All & Single-turn & Multi-turn & turns only \\
\midrule
Qwen3.5 vs.\ Llama3.1    & 50.0 & 53.4 {\scriptsize (51.0 -- 57.5)} & 50.5 {\scriptsize (49.2 -- 52.4)} & 54.6 {\scriptsize (50.5 -- 62.7)} & 99.2 {\scriptsize (98.3 -- 99.7)} \\
Qwen3.5 vs.\ Qwen2.5     & 50.0 & 54.1 {\scriptsize (50.7 -- 59.6)} & 50.7 {\scriptsize (49.4 -- 52.0)} & 55.4 {\scriptsize (50.9 -- 63.8)} & 98.8 {\scriptsize (97.8 -- 99.6)} \\
Qwen2.5 vs.\ Llama3.1    & 50.0 & 50.8 {\scriptsize (49.8 -- 53.0)} & 50.1 {\scriptsize (49.4 -- 51.1)} & 51.3 {\scriptsize (50.0 -- 54.9)} & 98.8 {\scriptsize (97.8 -- 99.6)} \\
Three-way                & 33.3 & 35.9 {\scriptsize (33.6 -- 38.8)} & 33.7 {\scriptsize (32.9 -- 34.6)} & 36.6 {\scriptsize (33.6 -- 42.1)} & 98.3 {\scriptsize (96.7 -- 99.4)} \\
\bottomrule
\end{tabular}
\end{table}

\vspace{10pt}

\textbf{Results.}
\autoref{tab:assistant_identity} shows the results.
Single-turn conversations are classified at chance (50.1\% -- 50.7\% on average in the binary comparisons, and 33.7\% in the three-way comparison),
as their only user message is generated before the assistant responds from the same user model.
Multi-turn conversations are classified above chance, averaging 51.3\% -- 55.4\% in the binary comparisons, and 36.6\% in the three-way comparison: the assistant leaves a trace in the user model's later turns.
This trace is weak compared with the assistant's own text, from which classifiers identify the assistant with 98.3\% -- 99.2\% accuracy.


\subsection{Cross-dataset Generalization}

\vspace{10pt}
\begin{table}[H]
\centering
\caption{Cross-dataset evaluation. Rows indicate training and columns indicate test (target) datasets. Each entry represents mean per-token perplexity measured on the test split. Bold and underlined entries indicate the lowest and second-lowest perplexity in each column, respectively.}
\label{tab:cross_dataset_evaluation}
\vspace{\figneg}
\vspace{\tablepos}
\small
\setlength{\tabcolsep}{4pt}
\renewcommand{\arraystretch}{1.15}
\resizebox{0.85\textwidth}{!}{%
\begin{tabular}{@{}l*{7}{r}@{}}
\toprule
    \shortstack[l]{Train $\downarrow$ Test $\rightarrow$}
    & WC-1M
    & WC-4.8M
    & LMSYS
    & Arena
    & ShareChat
    & ShareGPT
    & HH-RLHF \\
    \midrule
    WC-1M
    & \textbf{6.60} & \underline{7.05} & \underline{6.72} & 6.65 & 8.89
    & \underline{6.70} & 5.78 \\

    WC-4.8M
    & \underline{6.78} & \textbf{6.47} & 6.81 & \underline{6.54} & \underline{8.75}
    & 6.77 & 5.84 \\

    LMSYS
    & 6.93 & 7.30 & \textbf{5.60} & 6.71 & 8.90
    & 6.72 & \underline{5.60} \\

    Arena
    & 7.03 & 7.33 & 6.91 & \textbf{6.10} & 8.93
    & 6.93 & 5.76 \\

    ShareChat
    & 7.10 & 7.36 & 6.94 & 6.71 & \textbf{7.70}
    & 6.85 & 5.83 \\

    ShareGPT
    & 7.04 & 7.44 & 6.79 & 6.82 & 8.90
    & \textbf{5.93} & 5.62 \\

    HH-RLHF
    & 7.47 & 8.05 & 7.69 & 7.62 & 10.03
    & 7.40 & \textbf{4.86} \\
    \bottomrule
\end{tabular}%
}
\end{table}

\autoref{tab:cross_dataset_evaluation} reports the cross-dataset generalization results for user models based on Llama-3-8B.


\subsection{Classifier-guided Data Selection}

\begin{figure}[H]
    \centering
    \includegraphics[width=1.0\linewidth]{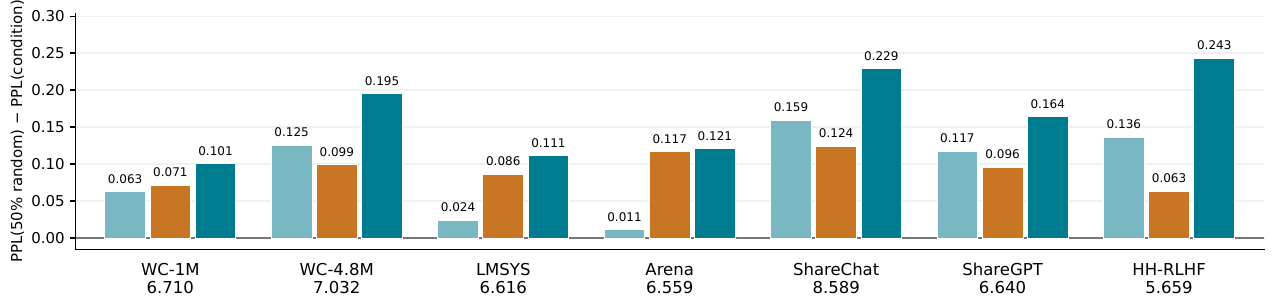}
    \vspace{\figneg}
    \vspace{\figneg}
    \caption{
    Perplexity reduction relative to uniform random sampling at 50\% of the training budget,
    PPL(50\% random) - PPL(condition), for Llama-3-8B user models (higher is better).
    The number under each dataset is the 50\%-random perplexity.
    Classifier-guided selection outperforms random sampling at every matched budget.
    }
    \label{fig:classifier-guided-ppl-llama}
\end{figure}

\textbf{Distributional alignment in classifier scores correlates with performance gains.}
Section~\ref{sec:training:cross-dataset} suggests that differences captured by dataset classifiers are relevant to cross-dataset generalization. We now examine a complementary question: when classifier-guided selection reduces the discrepancy between donor and target data, do larger reductions accompany larger improvements in predictive fit? Both questions examine whether classifier-derived measures of distributional mismatch are informative about the downstream user model performance.

Adapting the Kullback-Leibler reduction metric of \citet{xie2023data}, let $\hat p_T$, $\hat p_R$, and $\hat p_G$ denote the distributions of classifier scores (Softmax probabilities) for target, randomly selected donor, and classifier-guided donor, respectively. For each target dataset and train budget, we compute

\vspace{\figneg}
\begin{equation}
\Delta_{\mathrm{KL}}
=
D_{\mathrm{KL}}(\hat p_T \Vert \hat p_R)
-
D_{\mathrm{KL}}(\hat p_T \Vert \hat p_G)
\label{eq:kl_reduction}
\end{equation}
\vspace{\figneg}

A positive $\Delta_{\mathrm{KL}}$ indicates that guided selection brings the donor score distribution closer to the target score distribution than random sampling. We compare this quantity with the downstream improvement $\Delta_{\mathrm{NLL}} = \mathrm{NLL}_R - \mathrm{NLL}_G$, where $\mathrm{NLL}_R$ and $\mathrm{NLL}_G$ are the NLLs of user models trained on randomly selected and classifier-selected donors, respectively.

\begin{figure}[H]
    \centering
    \includegraphics[width=1.00\linewidth]{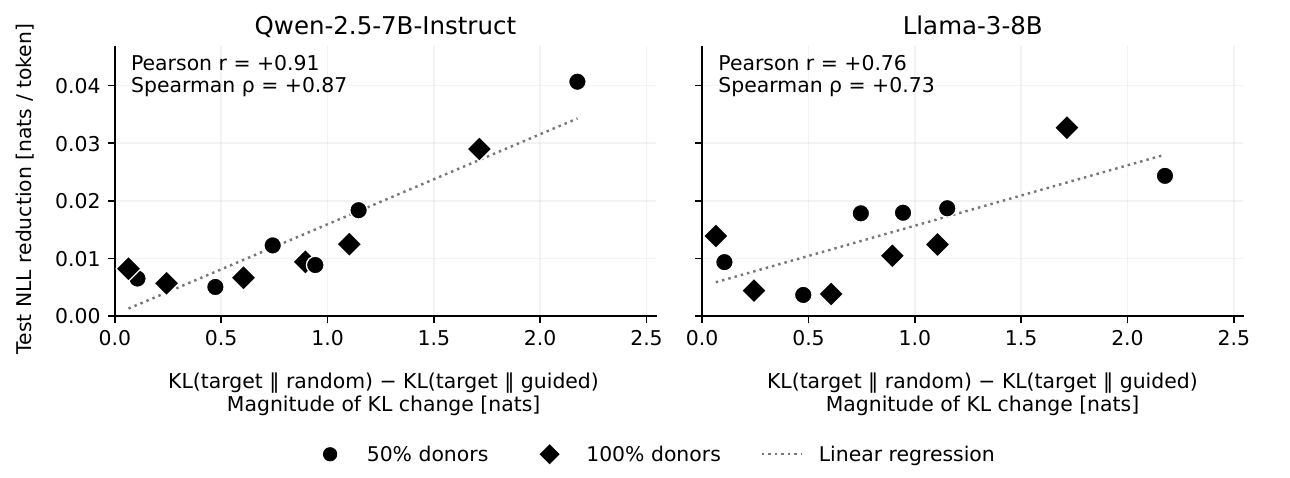}
    \vspace{\figneg}
    \vspace{\figneg}
    \caption{
    Magnitude of classifier-score distribution change versus downstream NLL improvement.
    The horizontal axis shows \autoref{eq:kl_reduction},
    computed from trained classifier scores for target,
    randomly selected donor, and classifier-selected donor.
    The vertical axis shows target NLL reduction, $\Delta_{\text{NLL}}$.
    }
    \label{fig:guidance-kl-nll}
\end{figure}

Figure~\ref{fig:guidance-kl-nll} shows a strong positive association between KL reduction and NLL improvement, with Pearson correlations of $r$=0.91 for Qwen-based and $r$=0.76 for Llama-based user models. This association suggests that the characteristics captured by the dataset classifier are relevant to selecting useful user-model training data.
We note that HH-RLHF target exhibits particularly large KL reductions and performance gains. One possible explanation is that its distinctive signatures, also reflected in its strong separability from other datasets (Section~\ref{sec:training}), provide a useful signal for identifying relevant conversations within the donor pool. Guided selection may therefore concentrate training on donors that share these signatures, yielding substantial improvements over uniform sampling.

\newpage
\end{document}